\documentclass[runningheads]{llncs}

\usepackage{iciap}

\usepackage{iciapabbrv}

\usepackage{graphicx}
\usepackage{booktabs}

\usepackage[accsupp]{axessibility}  % Improves PDF readability for those with disabilities.

\usepackage{adjustbox}
\usepackage{tabularray}
\usepackage{tabularx}    % For flexible tables
\usepackage{rotating}
\usepackage{pifont} % add in your preamble

\usepackage[pagebackref,breaklinks,colorlinks,citecolor=iciapblue]{hyperref}
\usepackage{orcidlink}
\begin{document}

\title{ReMAR-DS: Recalibrated Feature Learning for Metal Artifact Reduction and CT Domain Transformation} 

\titlerunning{ReMAR-DS}

\author{
Mubashara Rehman\inst{1,2}\orcidlink{0009-0007-2935-0409} \and
Niki Martinel\inst{1}\orcidlink{0000-0002-6962-8643} \and
Michele Avanzo\inst{2}\orcidlink{0000-0003-1711-4242} \and
Riccardo Spizzo\inst{2}\orcidlink{0000-0001-7772-0960} \and
Christian Micheloni\inst{1}\orcidlink{0000-0003-4503-7483}
}

\authorrunning{M. Rehman et al.}

\institute{
Università degli Studi di Udine, Udine, Italy \\
\email{rehman.mubashara@spes.uniud.it}\\
\email{\{niki.martinel, christian.micheloni\}@uniud.it} 
\and
Centro di Riferimento Oncologico di Aviano IRCCS, Aviano, Italy \\
\email{\{aumavanzo, rspizzo\}@cro.it}
}

\maketitle

\begin{abstract}
% Artifacts in kilo-Voltage Computed Tomography (kVCT) imaging degrade image quality, compromising clinical decision-making. This work proposes a deep learning-based framework for metal artifact reduction (MAR) and implicit domain transformation from kVCT to Mega-Voltage CT (MVCT). Unlike traditional MAR approaches, our method addresses domain transformation, a critical factor in integrating kVCT’s high-resolution imaging with MVCT’s artifact resistance, ensuring accurate radiotherapy planning in the presence of metal implants. The proposed framework ReMAR-DS, utilizes an encoder-decoder architecture with enhanced feature recalibration, effectively reducing artifacts while preserving anatomical structures. This ensures that only relevant information is utilized in the reconstruction process. By infusing recalibrated features from the encoder block, the model focuses on relevant spatial regions (e.g., areas with artifacts) and highlights key features across channels (e.g., anatomical structures), leading to improved reconstruction of artifact-corrupted regions. The method achieves high-quality MVCT reconstructions, demonstrating robust performance in both qualitative and quantitative evaluations. Clinically, this approach reduces radiation exposure for cancer patients by allowing radiation oncologists to derive insights from kVCT images alone, minimizing the need for repeated high-dose MVCT scans and reducing associated health risks.

% over limit >>> fit it to 150 words!!!

Artifacts in kilo-Voltage CT (kVCT) imaging degrade image quality, impacting clinical decisions. We propose a deep learning framework for metal artifact reduction (MAR) and domain transformation from kVCT to Mega-Voltage CT (MVCT). The proposed framework, ReMAR-DS, utilizes an encoder-decoder architecture with enhanced feature recalibration, effectively reducing artifacts while preserving anatomical structures. This ensures that only relevant information is utilized in the reconstruction process. By infusing recalibrated features from the encoder block, the model focuses on relevant spatial regions (e.g., areas with artifacts) and highlights key features across channels (e.g., anatomical structures), leading to improved reconstruction of artifact-corrupted regions. Unlike traditional MAR methods, our approach bridges the gap between high-resolution kVCT and artifact-resistant MVCT, enhancing radiotherapy planning. It produces high-quality MVCT-like reconstructions, validated through qualitative and quantitative evaluations. Clinically, this enables oncologists to rely on kVCT alone, reducing repeated high-dose MVCT scans and lowering radiation exposure for cancer patients.

\keywords{kilo-Voltage Computed Tomography \and Mega-Voltage Computed Tomography \and Metal artifact reduction }
\end{abstract}

\section{Introduction}
\label{sec:intro}

Computed tomography (CT) is a cornerstone in modern clinical diagnostics. However, the presence of metallic implants in patients poses significant challenges during the CT imaging process. These implants cause substantial X-ray attenuation, leading to incomplete X-ray projections and subsequently introducing severe bright and dark streaking \cite{Katsura2018,Lin2019,wang2021dicdnet,Liao2019} and shading artifacts in the reconstructed images. Such artifacts can critically impair the accuracy of clinical assessments and treatment planning \cite{Wellenberg2018}.

In recent years, various methods for metal artifact reduction (MAR) have been developed. Traditional approaches often replaced metal-corrupted regions in the sinogram domain with surrogate data, estimated through techniques like linear interpolation(LI) \cite{Kalender1987} and normalized MAR \cite{Meyer2010}. With the rise of deep learning, some methods have used neural networks to directly reconstruct clean sinograms  \cite{Ghani2019,Gjesteby2017,Liao2019ADN,Park2018,Zhang2018}. However, these approaches can introduce secondary artifacts due to inconsistencies with the physical imaging geometry and are limited by the difficulty of obtaining sinogram data in practice \cite{Liao2019}. To overcome these limitations, researchers have focused on training deep networks solely on CT images to recover artifact-free scans \cite{Gjesteby2018,Huang2018,Wang2018}. More recently, hybrid approaches that utilize both sinogram and CT data have been proposed, achieving improved MAR performance \cite{8953298,lyu2020encoding,Wang2021InDuDoNet,WANG2023102729,Wang2021c,9201079,ZHOU2022102289}. Nonetheless, many of these methods employ generic image restoration networks that do not fully leverage the specific characteristics of the MAR task, leading to challenges in network interpretability and effectiveness.
Our research not only addresses the challenge of MAR but also tackles the critical issue of domain transformation from kilo-Voltage Computed Tomography(kVCT) to Mega-Voltage Computed Tomography(MVCT). This dual focus ensures that our approach enhances image quality while facilitating accurate Radiotherapy(RT) planning in the presence of metal implants. kVCT, operating at 80-140 kVp, is the standard for diagnostic imaging due to its high spatial resolution and contrast sensitivity \cite{michielsen2013novel}. However, kVCT is highly susceptible to metal artifacts, particularly in patients with implants, which can obscure anatomical structures and compromise image quality. In contrast, MVCT, used in RT planning, operates at 1-10 MV and is less prone to metal artifacts due to reduced beam hardening \cite{maerz2015megavoltage}. MVCT offers more accurate electron density information critical for dose calculations \cite{duck2015evaluation}, but it suffers from lower spatial resolution, limiting its ability to visualize fine anatomical details. Recent advances in RT have highlighted the value of cross-modality image conversion, including MRI-to-CT~\cite{Wolterink2017,Maspero2018}, cone-beam computed tomography (CBCT) to CT for image-guided radiotherapy(IGRT) and adaptive radiotherapy(ART)~\cite{Kida2020,Liang_2019,Taasti2020}. In these conversions, domain-specific features from high-quality modalities are effectively integrated with those of the target domain, enabling the reconstruction of images that leverage the strengths of both and enhance overall quality.

In this research, we propose a domain transformation framework with implicit MAR, named as \textbf{ReMAR-DS}, in this model we have used enhanced residual blocks, that leverage depthwise separable convolutions to efficiently and effectively transform the encoded features, preparing them for the subsequent decoding phase while maintaining a high level of feature integrity and robustness. This model uses concurrent spatial and channel squeeze-and-excitation within the skip connections - inspired from research~\cite{roy2018concurrent}, which dynamically recalibrate features both spatially and across channels, ensuring the network focuses on the reconstruction of the regions affected by metal artifacts. This concurrent recalibration approach makes RscSE Blocks an essential part of our model for improving image quality in our cross-modality conversion.

Compared to existing MAR methods, our model shows consistent improvements in PSNR and SSIM on artifact-affected slices, confirming its robustness through clinical evaluation. It generates high-quality MVCT images from kVCT inputs, supporting better RT planning. Notably, by reducing the need for repeated high-dose imaging, it holds promise for minimizing patient radiation exposure.

% --------------------------------  
\begin{figure}[tb]
  \centering
  \begin{subfigure}[t]{0.45\linewidth}
    \centering
    \includegraphics[width=\linewidth]{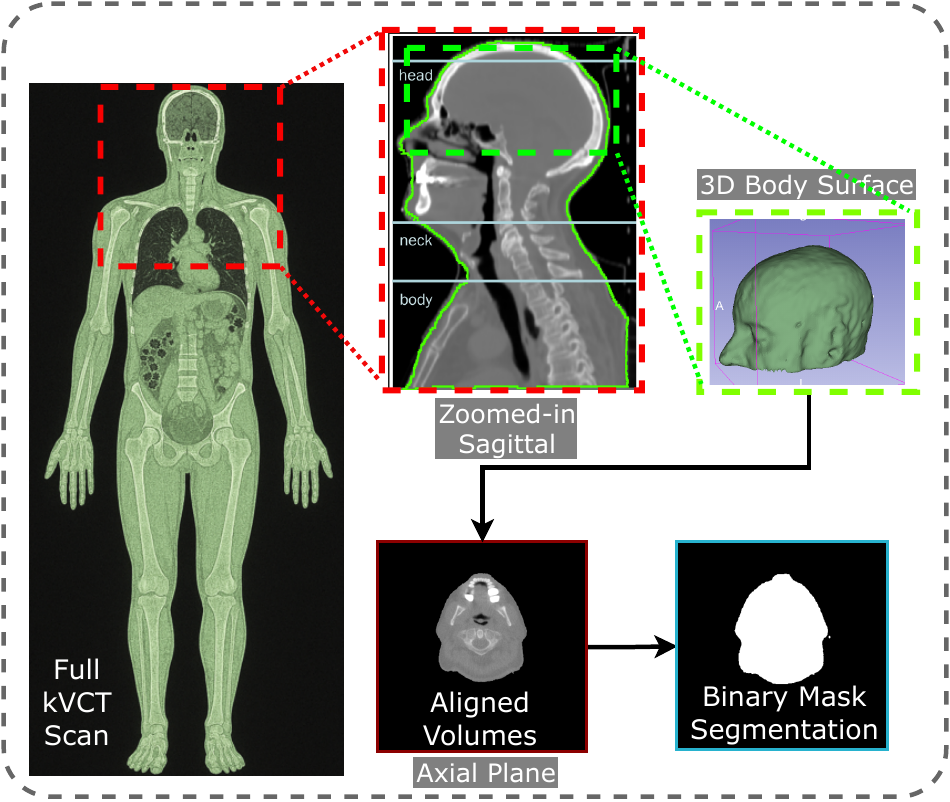}
    \caption{Clinician-assisted body segmentation masks}
    \label{fig:AlignedVolumes-maskExtraction}
  \end{subfigure}
  \hspace{0.03\linewidth} % <<-- Controlled spacing instead of \hfill
  \begin{subfigure}[t]{0.27\linewidth}
    \centering
    \includegraphics[width=0.9\linewidth]{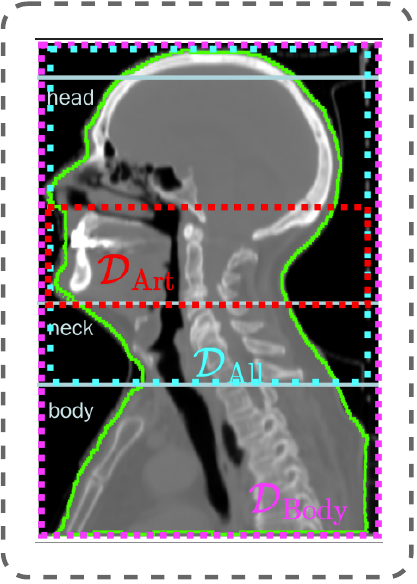}
    \caption{Dataset distribution used in experiments}
    \label{fig:regionsAndBodySegmenation}
  \end{subfigure}
  \vspace{-1mm}
  \caption{Overview of experimental setup and dataset distribution. (a) Clinician-assisted binary body masks are generated after aligning the kVCT and MVCT volumes. These masks are used during evaluation to exclude external artifacts and focus on the anatomical region of interest. (b) Visualization of dataset distribution: \( \mathcal{D}_\text{All} \) includes all slices, while \( \mathcal{D}_\text{Art} \) contains only artifact-contaminated slices.}
  \label{fig:exp-overview}
\end{figure}

% ========================================= Section
% =========================================

\section{Method}

\subsection{Problem Formulation}

The process of metal artifact removal and implicit domain transformation from kVCT to MVCT can be formulated as shown in \cref{eq:problem_formulation}

{\small
\begin{equation}
\mathbf{X}_k^{p} = \mathbf{X}_M^{recon} + A
\label{eq:problem_formulation}
\end{equation}
}

In this formulation, preprocessed kVCT images \(\mathbf{X}_k^{p}\) is the observation signal with dimensions \(\mathbb{R}^{512 \times 512 \times n}\) that has been corrupted by the artifacts \(A\), during the kVCT imaging process, while reconstructed images \(\mathbf{X}_M^{recon}\) is the target signal in the MVCT domain with dimensions \(\mathbb{R}^{512 \times 512 \times n}\). The artifact component \(A\) is present only in localized regions (e.g., the dental implant portion), and not across the entire \(\mathbf{X}_k^{p}\). The objective of the proposed framework is twofold:
\begin{enumerate}
    \item \textbf{Artifact Removal}: Estimating and removing the localized artifact component \(A\) from \(\mathbf{X}_k^{p}\).
    \item \textbf{Domain Transformation}: Mapping the resulting artifact-free kVCT image to the MVCT domain, resulting in \(\mathbf{X}_M^{recon}\).
\end{enumerate}

Thus, the task is to reconstruct \(\mathbf{X}_M^{recon}\) from the corrupted kVCT image \(\mathbf{X}_k^{p}\) by eliminating the artifacts \(A\) in specific regions and implicitly performing the domain transformation to the MVCT modality.

% ========================================= sub-Section
\subsection{Proposed Methodology}

\begin{figure*}[ht]
\centering
\includegraphics[width=0.99\textwidth]{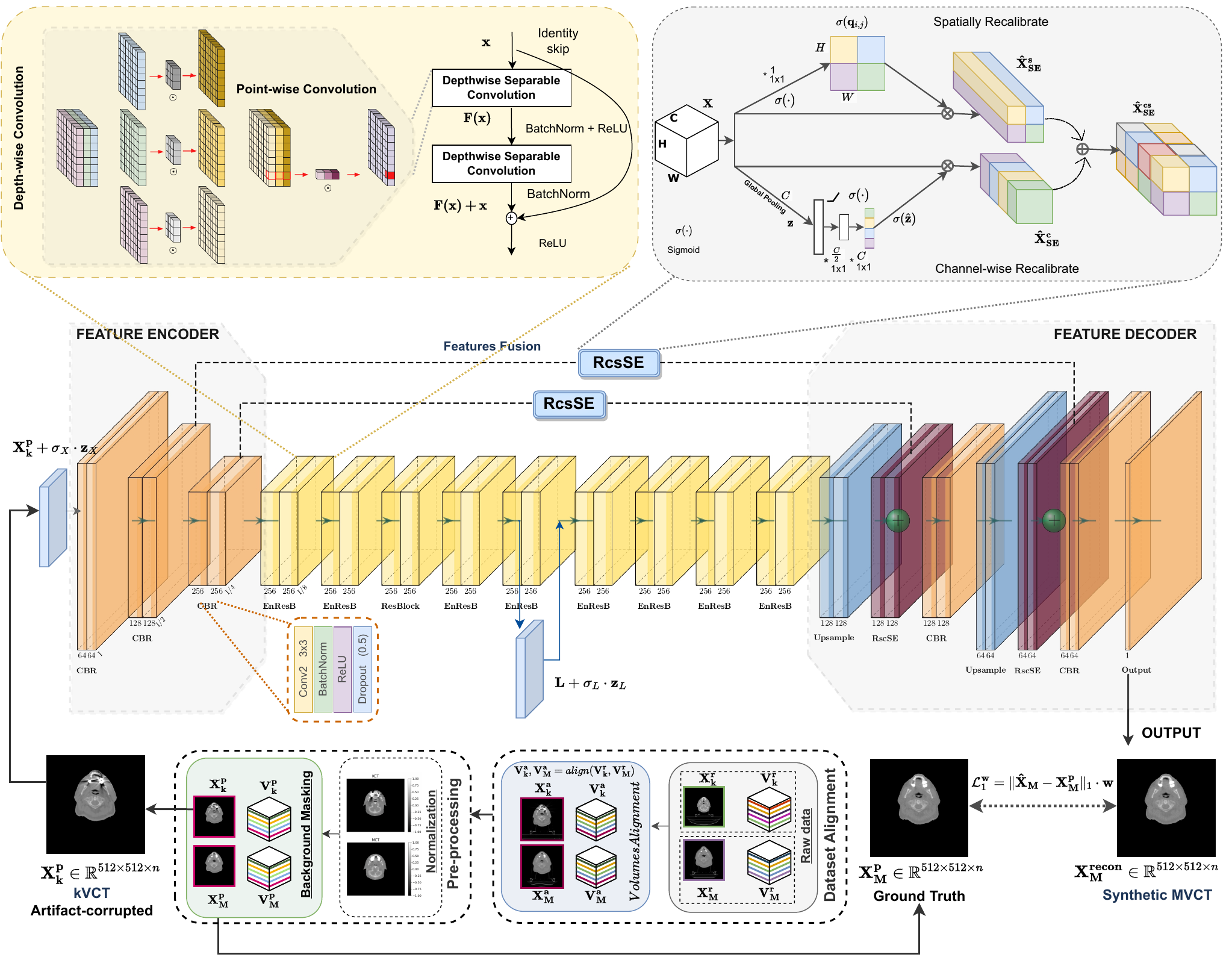}
\caption{Overview of the proposed model, ReMAR-DS for metal artifact reduction for kVCT to MVCT domain transformation, architecture successfully processes $512 \times 512$ pixel volumes to systematically mitigate metal artifacts.} 
\label{fig:overalarchitecture}
\end{figure*}

Our proposed model, \textbf{ReMAR-DS}, aims to reconstruct artifact-free MVCT slices \( \mathbf{X}_M^{\text{recon}} \in \mathbb{R}^{512 \times 512 \times n} \) from processed 2D kVCT inputs \( \mathbf{X}_k^p \in \mathbb{R}^{512 \times 512 \times n} \) affected by metal artifacts, as illustrated in \cref{fig:overalarchitecture}. During training, the input image \( \mathbf{X}_k^p \in D_{\text{dataset}}^{\text{Tr}} \) is perturbed with Gaussian noise \( z_k \sim \mathcal{N}(0,1) \), scaled by \( \sigma_k = 0.01 \), to improve model robustness and generalization. The encoded feature maps at each layer \( l \in \{1, \dots, L\} \) are computed as: $\mathbf{F}_l^{\text{enc}} = \text{Enc}_{kV}^l \left( \mathbf{X}_k^p + \sigma_k \cdot z_k \right)$, where \( \text{Enc}_{kV}^l(\cdot) \) denotes the encoder block at level \( l \), consisting of Conv-BN-ReLU (CBR) layers that progressively extract hierarchical features from low-level textures to high-level semantics. To improve representational efficiency, the encoder output is further transformed through Enhanced Residual Blocks (EnResB), where standard convolutions are replaced by depthwise separable convolutions. These blocks decouple spatial filtering and channel mixing via depthwise and pointwise operations, reducing the parameter count while enhancing sensitivity to spatial and channel-specific patterns. Residual connections within each block retain the original feature flow, supporting stable learning of deeper transformations.

To further enhance feature propagation, we introduce concurrent Spatial and Channel Squeeze-and-Excitation (RcsSE) blocks into the skip connections. For each layer \( l \), the encoder feature \( \mathbf{F}_l^{\text{enc}} \) is recalibrated to produce \( \mathbf{F}_l^{\text{RcsSE}} \), which is then added to the skip path. As illustrated in \cref{fig:overalarchitecture}, the spatial recalibration branch first applies a \( 1 \times 1 \) convolution to the input feature map \( \mathbf{X} \in \mathbb{R}^{H \times W \times C} \), generating a spatial attention map \( \mathbf{q} \in \mathbb{R}^{H \times W} \), followed by a sigmoid activation \( \sigma(\mathbf{q}_{i,j}) \) to emphasize informative spatial regions and suppress less relevant ones. In parallel, channel-wise recalibration is performed by global average pooling over \( \mathbf{X} \), followed by two fully connected layers and a sigmoid activation, yielding: $\mathbf{\hat{X}}_{\text{SE}}^{c} = \left[ \sigma(\hat{z}_1) \mathbf{x}_1, \dots, \sigma(\hat{z}_C) \mathbf{x}_C \right]
$
where \( \mathbf{x}_i \in \mathbb{R}^{H \times W} \) denotes the \( i \)-th channel in the set, for \( i = 1, \dots, C \), modulated by its learned importance \( \sigma(\hat{z}_i) \). By jointly recalibrating spatial and channel information, the RcsSE block allows the model to focus on artifact-prone regions while preserving modality-consistent anatomical structures across varying imaging conditions.

These recalibrated features \( \mathbf{F}_l^{\text{RcsSE}} \) are fused with the corresponding encoder outputs \( \mathbf{F}_l^{\text{enc}} \) and aggregated as input to the decoder \( \text{Dec}_{MV} \), which reconstructs the final synthetic MVCT image \( \mathbf{X}_M^{\text{recon}} \). The decoder comprises a series of upsampling and CBR layers that progressively refine the spatial resolution. Additionally, the deepest latent representation \( \mathbf{F}_L^{\text{enc}} \) is regularized with Gaussian noise \( \mathbf{z}_L \sim \mathcal{N}(0, 1) \), scaled by a learnable parameter \( \sigma_L \), to enhance robustness and generalization. The overall reconstruction process is defined in \cref{eq:model_overview_equation}.

{\small
\begin{equation}
\mathbf{X}_M^{\text{recon}} = \text{Dec}_{MV} \left( \sum_{l=1}^{L-1} \left( \mathbf{F}_l^{\text{enc}} + \mathbf{F}_l^{\text{RcsSE}} \right) + \left( \mathbf{F}_L^{\text{enc}} + \sigma_L \cdot z_L \right) \right),
\label{eq:model_overview_equation}
\end{equation}
}

This combination of noise-regularized encoding, efficient feature transformation, and attention-based skip recalibration enables ReMAR-DS to robustly suppress metal artifacts while preserving anatomical fidelity in the reconstructed MVCT output.

% -------------------------------- Figure
% -------------------------------- Figure

% ========================================= sub-Section

\subsection{Loss Functions}
\label{sub:loss_functions}

To assess the effectiveness of different objective functions in our reconstruction task, we conduct an ablation study using a range of complementary loss functions. These losses are combined in various configurations to analyze their individual and joint contributions to artifact suppression and perceptual fidelity.

The \textbf{weighted L1 Loss} (\( \mathcal{L}_1^w \)) introduces region-specific emphasis by assigning higher weights to clinically important areas, guided by body segmentation masks as shown in \cref{fig:AlignedVolumes-maskExtraction}. \textbf{SSIM} (\( \mathcal{L}_{\text{SSIM}} \)) and \textbf{Multi-Scale SSIM} (\( \mathcal{L}_{\text{MS-SSIM}} \)) prioritize structural coherence between the predicted and reference images across different spatial scales. To ensure numerical alignment at the pixel level, the standard \textbf{Mean Squared Error} (\( \mathcal{L}_{\text{MSE}} \)) is also included. Additionally, the \textbf{Focal Frequency Loss} (\( \mathcal{L}_{\text{FFL}}^{\beta, \alpha} \)) is utilized to enhance frequency-domain accuracy, particularly targeting high-frequency details often associated with sharp edges and fine textures.

By evaluating different combinations of these loss terms, we aim to identify the most effective formulation for achieving robust and high-quality MVCT reconstruction under varying artifact conditions.

{\small
\begin{align}
&\ \underbrace{\left( \lVert \mathbf{X_M^{recon}} -  \mathbf{X_M^p} \rVert_1 \cdot \mathbf{w} \right)}_{\mathcal{L}_1^{\textit{w}}}, \quad 
\underbrace{\left( 1 - \textit{SSIM}(\mathbf{X_M^{recon}}, \mathbf{X_M^p}) \right)}_{\mathcal{L}_{\textit{SSIM}}}, \quad
\underbrace{\left( 1 - \textit{MS-SSIM}(\mathbf{X_M^{recon}}, \mathbf{X_M^p}) \right)}_{\mathcal{L}_{\textit{MS-SSIM}}}, \nonumber \\ 
&\ \underbrace{\left( \lVert \mathbf{X_M^{recon}} -  \mathbf{X_M^p} \rVert_2^2 \right)}_{\mathcal{L}_{\textit{MSE}}}, \quad 
\underbrace{\left( \frac{1}{d \cdot d} \sum_{u=0}^{d-1}\sum_{v=0}^{d-1} z(u,v) \left| F_{\mathbf{X_M^{recon}}}(u,v) - F_{\mathbf{X_M^p}}(u,v) \right|^2 \cdot \beta \right)}_{\mathcal{L}_{\textit{FFL}}}
\label{eq:total_loss_with_commas}
\end{align}
}

% ============================================================= Experiments
\section{Experiments}
% --------------------------------  

\subsection{Dataset Acquisition and Distribution}
\label{sub:Dataset_Acquisition_and_Distribution}

% {National Cancer Institute (CRO) IRCCS}\footnote{Centro di Riferimento Oncologico di Aviano IRCCS, Via F. Gallini 2, Aviano (PN), 33081, Italy}

We utilize an unaligned dataset comprising kVCT and MVCT scans collected from the \textit{National Cancer Institute}\footnote{Centro di Riferimento Oncologico di Aviano IRCCS, Via F. Gallini 2, Aviano (PN), 33081, Italy}. The dataset includes imaging data from 52 patients who received intensity-modulated radiation therapy (IMRT) for oropharyngeal or nasopharyngeal cancer. The kVCT scans were acquired using a 32-slice CT scanner (Toshiba Aquilion LB), with a resolution of $512 \times 512$, 2\,mm slice thickness, and an in-plane pixel size of approximately 1.074\,mm. MVCT scans were captured using the Hi-Art II Tomotherapy System, featuring a 3.5\,MV beam and a matching resolution, but with a finer pixel size of 0.754\,mm and variable slice thickness (2–4\,mm).

To focus on regions affected by dental metal artifacts, each volume was segmented into three anatomical regions: head, neck, and upper torso (limited to the shoulder level) as shown in \cref{fig:regionsAndBodySegmenation}. Only the head and neck slices were retained for analysis. Artifact regions were identified using Hounsfield Unit (HU) thresholds: values above 2000\,HU in kVCT and above 1000\,HU in MVCT were considered artifact-prone, as guided by visual inspection and prior work~\cite{kim2022metal,liugang2016metal}. Based on this, we defined two datasets: $\mathcal{D}_\text{All}$, containing all head and neck slices, and $\mathcal{D}_\text{Art}$, comprising only slices with significant artifacts. Approximately 14.78\% of the total slices belong to $\mathcal{D}_\text{Art}$.

% --------------------------------  
\subsection{Implementation Details}
\label{sub:implementation_details}

Inputs and outputs are $512 \times 512$ slices ($\mathbf{X}_{k}^{p} \rightarrow \mathbf{X}_{M}^{\text{recon}}$). Training is conducted for 100 epochs with early stopping (patience 15) and batch size 4. AdamW is used with an initial learning rate of $1\mathrm{e}{-4}$ (halved every 20 epochs), weight decay $5e^{-4}$, and gradient clipping. Augmentation via Albumentations~\cite{Buslaev_2020} includes horizontal flip (0.5) and affine transforms (0.8 probability; shift=0.0625, scale=0.1, rotate=5). Each image yields 2–3 variants. Models run on an Intel Xeon CPU (188\,GB RAM) and NVIDIA A100. Datasets $\mathcal{D}_\text{All}$ and $\mathcal{D}_\text{Art}$ are split patient-wise: 70\% for training, 30\% for testing, forming $\mathcal{D}_\text{All}^{Tr/Ts}$ and $\mathcal{D}_\text{Art}^{Tr/Ts}$. Loss weights for $\mathcal{L}_1^{w}$ are 100 (inside body), 0.1 (outside) for artifact slices; 1 for clean slices. Focal frequency loss uses $\beta=1$, $\alpha=0.5$~\cite{mardtn-icpr2024}.

% --------------------------------------
\subsection{Baseline and Evaluation Metric}

In this study, we conduct a comprehensive evaluation of our proposed network's performance compared to several state-of-the-art methods. These include:  \textbf{MAR-DTN}\cite{mardtn-icpr2024}, \textbf{Pix2Pix}, a Conditional Generative Adversarial Network \cite{8100115},
\textbf{Mod-cGAN}, a modified version of Pix2Pix using the approach from \cite{PMID:30693351}, \textbf{SwinIR}, a Transformer-based architecture~\cite{liang2021swinir}, \textbf{AttU-Net}, and a modified Attention-based UNet \cite{wang2022attu-net}. For the comparative analysis, we report the results using the optimal loss function configurations (\cref{sub:loss_functions}) as established in our previous work. For quantitative assessment we adopt PSNR/SSIM; which are calculated only in the region of interest using the masks as shown in \cref{fig:AlignedVolumes-maskExtraction}. As the ground truth MVCT dataset is not entirely artifact-free, so due to unavailability of reference dataset for artifact-corrupted kVCT slices, clinician evaluation, and qualitative results are also provided.

% ============================== Result and Discussion
\section{Result and Discussion}
\subsection{Ablation Study}

Comparative evaluation of the ReMAR-DS model is presented in \cref{table:ablation_lossfunctions}, trained using various loss function combinations on both \(\mathcal{D}_\text{All}^{Tr}\) and \(\mathcal{D}_\text{Art}^{Tr}\), and tested on their respective test sets. The results highlight how different loss functions influence reconstruction quality, particularly in artifact-prone regions versus the overall dataset.

% -----------------  Ablation Table for Loss-Functions

\begin{table}
\centering
\caption{Comparative analysis of ReMAR-DS with different loss combinations and datasets.}
\label{table:ablation_lossfunctions}
\tiny
\begin{tblr}{
  row{2} = {c},
  row{3} = {c},
  row{4} = {c},
  row{5} = {c},
  row{7} = {c},
  row{8} = {c},
  row{9} = {c},
  cell{1}{1} = {r=5}{} ,
  cell{1}{2} = {c},
  cell{1}{3} = {c},
  cell{1}{4} = {c},
  cell{1}{5} = {c},
  cell{1}{6} = {c},
  cell{1}{7} = {c},
  cell{1}{8} = {c},
  cell{6}{1} = {r=4}{} ,
  cell{6}{2} = {c},
  cell{6}{3} = {c},
  cell{6}{4} = {c},
  cell{6}{5} = {c},
  cell{6}{6} = {c},
  cell{6}{7} = {c},
  cell{6}{8} = {c},
  vline{1} = {1-9}{},
  vline{9} = {1-9}{},
  hline{10} = {1-8}{},
  hline{6} = {-}{},
  hline{1} = {-}{},
  hline{8} = {2-8}{},
}
\begin{sideways}\textbf{\textbf{Loss Combination}}\end{sideways} & $\mathcal{L}_1^{100}$              & $\checkmark$               & $\checkmark$      & $\checkmark$      & $\checkmark$      & $\checkmark$      & $\checkmark$      \\
 & $\mathcal{L}_\textit{SSIM}$      & \ding{55}          & $\checkmark$      &   \ding{55}                &                 \ding{55}  &            \ding{55}       & $\checkmark$      \\
& $\mathcal{L}_\textit{MS-SSIM}$   &                 \ding{55}           &   \ding{55}                & $\checkmark$      &   \ding{55}                &  \ding{55}       &       \ding{55}            \\
& $\mathcal{L}_\textit{MSE}$       &      \ding{55}                      &      \ding{55}             &   \ding{55}                & $\checkmark$      &     \ding{55}              &  \ding{55}                 \\
& $\mathcal{L}_\textit{FFL}^{1,0.5}$ &                 \ding{55}           &   \ding{55}                &              \ding{55}     &         \ding{55}          & $\checkmark$      & $\checkmark$      \\
\begin{sideways}\textbf{\textbf{Dataset}}\end{sideways}          & $\mathcal{D}_\text{Art}$ (PSNR)  & $27.260$                   & $27.260$          & $27.280$          & $27.455$          & $27.230$          & $\textbf{27.670}$ \\
& $\mathcal{D}_\text{Art}$ (SSIM)  & $0.700$                    & $0.704$           & $0.704$           & $0.712$           & $0.704$           & $\textbf{0.703}$  \\
& $\mathcal{D}_\text{All}$ (PSNR)  & {$\textbf{27.852}$ \\ \textbf{(30.690)}} & {$28.142$ \\ (30.650)} & {$28.017$ \\ (30.487)} & {$27.925$ \\ (30.388)} & {$28.141$ \\ (30.511)} & {$27.921$ \\ (30.090)} \\
& $\mathcal{D}_\text{All}$ (SSIM)  & {$\textbf{0.716}$ \\ \textbf{(0.759)}} & {$0.726$ \\ (0.768)} & {$0.725$ \\ (0.765)} & {$0.720$ \\ (0.757)} & {$0.725$ \\ (0.761)} & {$0.721$ \\ (0.765)}
\end{tblr}
\end{table}

% ---------------------------------------------

On the full test dataset \(\mathcal{D}_\text{All}^{Ts}\), the best performance is observed when using the simple weighted \( \mathcal{L}_1^{100} \) loss alone. It achieves the highest PSNR of 30.690 and SSIM of 0.759, indicating that pixel-wise fidelity is sufficient when evaluating both clean and artifact-affected slices collectively. This suggests that \( \mathcal{L}_1 \) provides strong generalization and intensity consistency across the whole dataset. However, when focusing specifically on the artifact-heavy subset \(\mathcal{D}_\text{Art}^{Ts}\), the loss combination \( \mathcal{L}_1^{100} + \mathcal{L}_{\textit{SSIM}} + \mathcal{L}_{\textit{FFL}}^{1,0.5} \) delivers superior performance, achieving the highest PSNR of 27.670. This shows that structural similarity and frequency-aware refinement play a critical role in restoring images with severe metal-induced degradation.

The ablation reveals that while \( \mathcal{L}_1^{100} \) excels on the overall dataset, integrating structural and frequency-domain losses significantly boosts performance on challenging slices. Thus, a hybrid formulation with \( \mathcal{L}_\textit{SSIM} \) and \( \mathcal{L}_\textit{FFL} \) is essential for improving reconstructions in artifact-prone regions, while preserving fidelity in fine details.

% grid plot----------------------------
\begin{figure}[h!]
\centering
\includegraphics[width=0.7\textwidth]{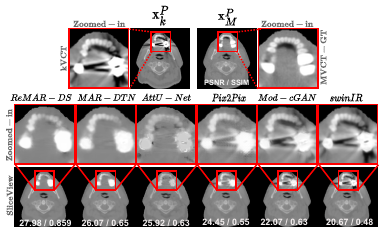}
\caption{Reconstruction of an artifact-corrupted MVCT slice using different models. The first row shows the preprocessed kVCT and MVCT (ground truth). All networks are trained using \(D_{\text{Art}}^{Tr}\), with PSNR and SSIM metrics provided for each reconstruction. Display window: [-1000, 1000] HU.}
\label{fig:sliceReconstruction}
\end{figure}
% ---------------------------------------------

\subsection{State-of-the-Art Model Comparison}
\label{subsec:sota_comparison}

A comprehensive quantitative comparison between the proposed \textbf{ReMAR-DS} model and several state-of-the-art (SOTA) baselines is presented in \cref{table:comparativeResults_sota}. The evaluation is conducted on both artifact-affected slices ($\mathcal{D}_\text{Art}^{Ts}$) and the complete test set ($\mathcal{D}_\text{All}^{Ts}$). Our proposed model achieves the highest performance across all metrics and datasets, with a PSNR of 27.670 dB and SSIM of 0.703 on \(\mathcal{D}_\text{Art}^{Ts}\), and 30.690 dB and 0.759 on \(\mathcal{D}_\text{All}^{Ts}\). This demonstrates its superior generalization to both artifact-heavy and mixed-quality data.

The table also includes four ablation variants of ReMAR-DS to assess the contribution of key architectural components. \textbf{ReMAR-DS\textsuperscript{+}} removes the RcsSE block from skip connections; \textbf{ReMAR-DS\textsuperscript{++}} excludes Gaussian noise; \textbf{ReMAR-DS\textsuperscript{+++}} omits both Gaussian noise and RcsSE; and \textbf{ReMAR-DS\textsuperscript{++++}} further replaces enhanced residual blocks (EnResB) with simpler residual blocks. 

As expected, performance gradually declines with the removal of each component, confirming the importance of spatial recalibration, noise regularization, and enriched residual learning. Compared to other SOTA models, including MAR-DTN~\cite{mardtn-icpr2024}, AttU-Net~\cite{wang2022attu-net}, Pix2Pix~\cite{8100115}, and Mod-cGAN~\cite{PMID:30693351}, ReMAR-DS consistently outperforms in both PSNR and SSIM. Notably, SwinIR~\cite{liang2021swinir}, although designed for super-resolution, shows limited effectiveness in artifact reduction, with the lowest SSIM values across both datasets.

Overall, the results validate the robustness and effectiveness of ReMAR-DS in MAR and domain transformation, particularly in clinically significant regions affected by dental implants. A visual comparison of reconstruction quality across models is illustrated in \cref{fig:sliceReconstruction}, further demonstrating the effectiveness of ReMAR-DS in recovering artifact-corrupted regions.

% ------------------------ SOTA table

\begin{table}[ht]
\centering
\tiny
\caption{Quantitative comparison of ReMAR-DS and baseline models on $\mathcal{D}_\text{Art}^{Ts}$ and $\mathcal{D}_\text{All}^{Ts}$. Ablation variants of ReMAR-DS highlight the contribution of key components. Bold values indicate the best performance. PSNR and SSIM are computed within regions of interest (ROI), as defined by clinician-provided anatomical segmentations, specifically around metal-affected areas.}
\label{table:comparativeResults_sota}
\renewcommand{\arraystretch}{1.5} % Adjust row height

\begin{tabular}{|c|c|c|c|c|}
\hline
\textbf{Models}                  & \multicolumn{2}{c|}{$\mathcal{D}_\text{Art}$} & \multicolumn{2}{c|}{$\mathcal{D}_\text{All}$} \\  

% \hline
& \textbf{PSNR} & \textbf{SSIM} & \textbf{PSNR} & \textbf{SSIM} \\ 
\hline
$\mathbf{ReMAR}$-$\mathbf{DS}$ (Ours)   & \textbf{27.670}        & \textbf{0.703}         & \textbf{27.852 (30.690)} & \textbf{0.716 (0.759)} \\ 
$\mathbf{ReMAR}$-$\mathbf{DS}\textsuperscript{+}$     & 27.321          & 0.695          & 27.496 (30.410)          & 0.709 (0.751) \\
$\mathbf{ReMAR}$-$\mathbf{DS}\textsuperscript{++}$    & 27.258          & 0.690          & 27.385 (30.322)          & 0.705 (0.747) \\
$\mathbf{ReMAR}$-$\mathbf{DS}\textsuperscript{+++}$   & 26.874          & 0.673          & 26.963 (29.924)          & 0.692 (0.734) \\
$\mathbf{ReMAR}$-$\mathbf{DS}\textsuperscript{++++}$  & 26.342          & 0.645          & 26.421 (29.201)          & 0.664 (0.715) \\
\hline

$\mathbf{MAR}$-$\mathbf{DTN}$ \cite{mardtn-icpr2024} & 27.46         & 0.69          & 27.05 (30.02)  & 0.69 (0.73)  \\ 
% \hline
$\mathbf{AttU}$-$\mathbf{Net}$ \cite{wang2022attu-net} & 27.09         & 0.69          & 26.69 (30.12) & 0.68 (0.73)  \\ 
% \hline
$\mathbf{Pix}$2$\mathbf{Pix}$ \cite{8100115}    & 26.31         & 0.64          & 26.36 (28.7)  & 0.63 (0.69)  \\ 
% \hline
$\mathbf{Mod}$-$\mathbf{cGAN}$ \cite{PMID:30693351}   & 25.24         & 0.68          & 26.61 (29.08) & 0.67 (0.7)   \\ 
% \hline
$\mathbf{swinIR}$ \cite{liang2021swinir}       & 26.18         & 0.63          & 25.24 (29.00) & 0.60 (0.66)  \\ \hline
\end{tabular}
\end{table}

% -------------------------------------------

% ============================================================= Conclusion
\section{Conclusion and Future Work}

We proposed a deep learning framework for metal artifact reduction and domain transformation from kVCT to MVCT. By leveraging a recalibrated encoder-decoder design, the model effectively mitigates artifacts while preserving anatomical integrity. Quantitative gains in PSNR and SSIM, supported by clinical feedback, highlight its potential to enhance image quality and reduce repeat scans. Future work will focus on improving contrast consistency, refining HU accuracy, and extending the method to 3D volumes for broader clinical applicability.

% =========================================================
\subsubsection*{Acknowledgments}
This work was supported by the Italian Ministry of Health (Ricerca Corrente 2024). We thank the Centro di Riferimento Oncologico di Aviano IRCCS, for providing the datasets and resources used in this study.

% ================================================

\bibliographystyle{splncs04}
\bibliography{main}

\begin{thebibliography}{10}
\providecommand{\url}[1]{\texttt{#1}}
\providecommand{\urlprefix}{URL }
\providecommand{\doi}[1]{https://doi.org/#1}

\bibitem{Buslaev_2020}
Buslaev, A., Iglovikov, V.I., Khvedchenya, E., Parinov, A., Druzhinin, M., Kalinin, A.A.: Albumentations: Fast and flexible image augmentations. Information  \textbf{11}(2), ~125 (Feb 2020)

\bibitem{duck2015evaluation}
Duck, L., McEwen, M., Wong, T., Brown, C.: Evaluation of metal artifact reduction algorithms for ct imaging in radiotherapy planning. Radiotherapy and Oncology  \textbf{117}(3),  460--464 (2015)

\bibitem{Ghani2019}
Ghani, M.U., Karl, W.C.: Fast enhanced ct metal artifact reduction using data domain deep learning. IEEE Transactions on Computational Imaging  \textbf{5}(4),  609--619 (2019). \doi{10.1109/TCI.2019.2929372}

\bibitem{Gjesteby2017}
Gjesteby, L., Yang, Q., Xi, Y., Zhou, Y., Zhang, J., Wang, G.: Deep learning methods to guide ct image reconstruction and reduce metal artifacts. In: Proceedings of the SPIE Conference on Medical Imaging: Physics of Medical Imaging. vol. 10132, p. 101322W (2017). \doi{10.1117/12.2256950}

\bibitem{Gjesteby2018}
Gjesteby, L., et~al.: Deep neural network for ct metal artifact reduction with a perceptual loss function. In: Proceedings of the Fifth International Conference on Image Formation in X-ray Computed Tomography. pp. 430--434 (2018)

\bibitem{Huang2018}
Huang, X., Wang, J., Tang, F., Zhong, T., Zhang, Y.: Metal artifact reduction on cervical ct images by deep residual learning. Biomedical Engineering Online  \textbf{17}(1),  1--20 (2018). \doi{10.1186/s12938-018-0471-6}

\bibitem{8100115}
Isola, P., Zhu, J.Y., Zhou, T., Efros, A.A.: Image-to-image translation with conditional adversarial networks. In: 2017 IEEE Conference on Computer Vision and Pattern Recognition (CVPR). pp. 5967--5976 (2017)

\bibitem{Kalender1987}
Kalender, W.A., Hebel, R., Ebersberger, J.: Reduction of ct artifacts caused by metallic implants. Radiology  \textbf{164}(2),  576--577 (1987)

\bibitem{Katsura2018}
Katsura, M., Sato, J., Akahane, M., Kunimatsu, A., Abe, O.: Current and novel techniques for metal artifact reduction at ct: practical guide for radiologists. Radiographics  \textbf{38}(2),  450--461 (2018)

\bibitem{Kida2020}
Kida, S., Kaji, S., Nawa, K., et~al.: Visual enhancement of cone-beam ct by use of cyclegan. Medical Physics  \textbf{47}(3),  998--1010 (2020). \doi{10.1002/mp.14010}

\bibitem{kim2022metal}
Kim, H., Yoo, S.K., Kim, D.W., Lee, H., Hong, C.S., Han, M.C., Kim, J.S.: Metal artifact reduction in kv ct images throughout two-step sequential deep convolutional neural networks by combining multi-modal imaging (martian). Scientific Reports  \textbf{12}(1),  20823 (2022)

\bibitem{liang2021swinir}
Liang, J., Cao, J., Sun, G., Zhang, K., Van~Gool, L., Timofte, R.: Swinir: Image restoration using swin transformer. In: Proceedings of the IEEE/CVF international conference on computer vision. pp. 1833--1844 (2021)

\bibitem{Liang_2019}
Liang, X., Chen, L., Nguyen, D., Zhou, Z., Gu, X., Yang, M., Wang, J., Jiang, S.: Generating synthesized computed tomography (ct) from cone-beam computed tomography (cbct) using cyclegan for adaptive radiation therapy. Physics in Medicine \& Biology  \textbf{64}(12),  125002 (jun 2019)

\bibitem{Liao2019}
Liao, H., Lin, W.A., Zhou, S.K., Luo, J.: Adn: artifact disentanglement network for unsupervised metal artifact reduction. IEEE Transactions on Medical Imaging  \textbf{38}(9),  2119--2130 (2019). \doi{10.1109/TMI.2019.2905891}

\bibitem{Liao2019ADN}
Liao, H., Lin, W.A., Zhou, S.K., Luo, J.: {ADN}: artifact disentanglement network for unsupervised metal artifact reduction. IEEE Transactions on Medical Imaging  \textbf{38}(1),  131--141 (2019). \doi{10.1109/TMI.2018.2868013}

\bibitem{8953298}
Lin, W.A., Liao, H., Peng, C., Sun, X., Zhang, J., Luo, J., Chellappa, R., Zhou, S.K.: Dudonet: Dual domain network for ct metal artifact reduction. In: 2019 IEEE/CVF Conference on Computer Vision and Pattern Recognition (CVPR). pp. 10504--10513 (2019)

\bibitem{Lin2019}
Lin, W.A., Liao, H., Zhou, S.K., Luo, J.: Dudonet: dual domain network for ct metal artifact reduction. In: Proceedings of the IEEE/CVF Conference on Computer Vision and Pattern Recognition (CVPR). pp. 10512--10521 (2019)

\bibitem{liugang2016metal}
Liugang, G., Hongfei, S., Xinye, N., Mingming, F., Zheng, C., Tao, L.: Metal artifact reduction through mvcbct and kvct in radiotherapy. Scientific reports  \textbf{6}(1),  37608 (2016)

\bibitem{lyu2020encoding}
Lyu, Y., Lin, W., Liao, H., Lu, J., Zhou, S.: Encoding metal mask projection for metal artifact reduction in computed tomography. In: Medical Image Computing and Computer Assisted Intervention – MICCAI 2020: 23rd International Conference, Lima, Peru, October 4–8, 2020, Proceedings, Part II. pp. 147--157. Springer (2020)

\bibitem{maerz2015megavoltage}
Maerz, M., Johnson, S., Cox, J.: Megavoltage computed tomography imaging: a review of current technology and applications in radiation therapy. Journal of Medical Imaging and Radiation Sciences  \textbf{46}(3),  291--297 (2015)

\bibitem{Maspero2018}
Maspero, M., Savenije, M.H., Dinkla, A.M., et~al.: Dose evaluation of fast synthetic-ct generation using a generative adversarial network for general pelvis mr-only radiotherapy. Physics in Medicine \& Biology  \textbf{63}(18),  185001 (2018)

\bibitem{Meyer2010}
Meyer, E., Raupach, R., Lell, M., Schmidt, B., Kachelrie{\ss}, M.: Normalized metal artifact reduction (nmar) in computed tomography. Medical Physics  \textbf{37}(10),  5482--5493 (2010). \doi{10.1118/1.3488989}

\bibitem{michielsen2013novel}
Michielsen, K., Xia, T., Boone, J.: A novel metal artifact reduction algorithm in ct imaging for the improvement of radiotherapy planning. Medical Physics  \textbf{40}(8),  081917 (2013)

\bibitem{Park2018}
Park, H.S., Lee, S.M., Kim, H.P., Seo, J.K., Chung, Y.E.: Ct sinogram-consistency learning for metal-induced beam hardening correction. Medical Physics  \textbf{45}(12),  5376--5384 (2018). \doi{10.1002/mp.13208}

\bibitem{roy2018concurrent}
Roy, A.G., Navab, N., Wachinger, C.: Concurrent spatial and channel ‘squeeze \& excitation’in fully convolutional networks. In: International conference on medical image computing and computer-assisted intervention. pp. 421--429. Springer (2018)

\bibitem{mardtn-icpr2024}
Serrano-Ant{\'o}n, B., Rehman, M., Martinel, N., Avanzo, M., Spizzo, R., Fanetti, G., P.~Mu{\~{n}}uzuri, A., Micheloni, C.: Mar-dtn: Metal artifact reduction using domain transformation network for radiotherapy planning. In: Pattern Recognition. pp. 143--159. Springer Nature Switzerland (2025)

\bibitem{Taasti2020}
Taasti, V.T., Klages, P., Parodi, K., Muren, L.P.: Developments in deep learning based corrections of cone beam computed tomography to enable dose calculations for adaptive radiotherapy. Physics and Imaging in Radiation Oncology  \textbf{15},  77--79 (2020)

\bibitem{wang2021dicdnet}
Wang, H., Li, Y., He, N., Ma, K., Meng, D., Zheng, Y.: Dicdnet: deep interpretable convolutional dictionary network for metal artifact reduction in ct images. IEEE Transactions on Medical Imaging  \textbf{41}(4),  869--880 (2021)

\bibitem{Wang2021InDuDoNet}
Wang, H., et~al.: Indudonet: An interpretable dual domain network for ct metal artifact reduction. In: International Conference on Medical Image Computing and Computer-Assisted Intervention (MICCAI). pp. 107--118. Springer, Cham (2021). \doi{10.1007/978-3-030-87196-3_11}

\bibitem{WANG2023102729}
Wang, H., Li, Y., Zhang, H., Meng, D., Zheng, Y.: Indudonet+: A deep unfolding dual domain network for metal artifact reduction in ct images. Medical Image Analysis  \textbf{85},  102729 (2023)

\bibitem{Wang2018}
Wang, J., Zhao, Y., Noble, J.H., Dawant, B.M.: Conditional generative adversarial networks for metal artifact reduction in ct images of the ear. In: International Conference on Medical Image Computing and Computer-Assisted Intervention (MICCAI). pp. 3--11. Springer, Cham (2018). \doi{10.1007/978-3-030-00934-2_1}

\bibitem{PMID:30693351}
Wang, J., Zhao, Y., Noble, J.H., Dawant, B.M.: Conditional generative adversarial networks for metal artifact reduction in ct images of the ear. Medical image computing and computer-assisted intervention : MICCAI ... International Conference on Medical Image Computing and Computer-Assisted Intervention  \textbf{11070},  3—11 (September 2018). \doi{10.1007/978-3-030-00928-1_1}, \url{https://europepmc.org/articles/PMC6347117}

\bibitem{wang2022attu-net}
Wang, S., Li, L., Zhuang, X.: {AttU-NET}: Attention {U-Net} for brain tumor segmentation. In: Crimi, A., Bakas, S. (eds.) Brainlesion: Glioma, Multiple Sclerosis, Stroke and Traumatic Brain Injuries, Lecture Notes in Computer Science, vol. 13218, pp. 302--311. Springer International Publishing, Cham (2022). \doi{10.1007/978-3-031-09002-8_27}

\bibitem{Wang2021c}
Wang, T., et~al.: Dual-domain adaptive-scaling non-local network for ct metal artifact reduction. In: International Conference on Medical Image Computing and Computer-Assisted Intervention (MICCAI). Lecture Notes in Computer Science (LNCS), vol. 12906, pp. 243--253. Springer, Cham (2021). \doi{10.1007/978-3-030-87231-1_24}

\bibitem{Wellenberg2018}
Wellenberg, R.H.H., Hakvoort, E.T., Slump, C.H., Boomsma, M.F., Maas, M., Streekstra, G.J.: Metal artifact reduction techniques in musculoskeletal ct-imaging. European Journal of Radiology  \textbf{107},  60--69 (2018)

\bibitem{Wolterink2017}
Wolterink, J.M., Dinkla, A.M., Savenije, M.H., Seevinck, P.R., van~den Berg, C.A., I{\v{s}}gum, I.: Deep mr to ct synthesis using unpaired data. In: International Workshop on Simulation and Synthesis in Medical Imaging. pp. 14--23. Springer, Cham (2017)

\bibitem{9201079}
Yu, L., Zhang, Z., Li, X., Xing, L.: Deep sinogram completion with image prior for metal artifact reduction in ct images. IEEE Transactions on Medical Imaging  \textbf{40}(1),  228--238 (2021)

\bibitem{Zhang2018}
Zhang, Y., Yu, H.: Convolutional neural network based metal artifact reduction in x-ray computed tomography. IEEE Transactions on Medical Imaging  \textbf{37}(6),  1370--1381 (2018). \doi{10.1109/TMI.2018.2820342}

\bibitem{ZHOU2022102289}
Zhou, B., Chen, X., Zhou, S.K., Duncan, J.S., Liu, C.: Dudodr-net: Dual-domain data consistent recurrent network for simultaneous sparse view and metal artifact reduction in computed tomography. Medical Image Analysis  \textbf{75},  102289 (2022)

\end{thebibliography}
\end{document}